\documentclass{article}
\usepackage{spconf,amsmath,graphicx}
\usepackage{xcolor}

\usepackage{tikz}
\usepackage{graphicx}
\usetikzlibrary{positioning}
\usepackage[font=small]{caption}
\usepackage{amsfonts}

\title{MULTI-LEVEL NETWORK FOR HIGH-SPEED MULTI-PERSON POSE ESTIMATION}
\name{Ying Huang$^\dagger{}^*$   \quad Jiankai Zhuang $^\ddagger{}^*$\quad Zengchang Qin$^\ddagger{}^*$\thanks{$^*$The work was done at Keep Inc. The research was partially supported by the National Key Research and Development Program of China (2017YFB1002803).}}
\address{$^\dagger$ Alibaba Business School, Hangzhou Normal University, Hangzhou, China \\
$^\ddagger$ Intelligent Computing and Machine Learning Lab, School of ASEE, Beihang University, Beijing, China 
\\
\small \texttt{\{yw155, zhuangjk, zcqin\}@buaa.edu.cn}
}

\begin{document}
%\ninept
%
\maketitle
\begin{abstract}
In multi-person pose estimation, the left/right joint type discrimination is always a hard problem because of the similar appearance. Traditionally, we solve this problem by stacking multiple refinement modules to increase network's receptive fields and capture more global context, which can also increase a great amount of computation. In this paper, we propose a Multi-level Network (MLN) that learns to aggregate features from lower-level (left/right information), upper-level (localization information), joint-limb level (complementary information) and global-level (context) information for discrimination of joint type. Through feature reuse and its intra-relation, MLN can attain comparable performance to other conventional methods while runtime speed retains at 42 FPS.
\end{abstract}

\section{Introduction}
\label{sec:intro}

Human pose estimation is a basic problem in computer vision which aims to extract the posture of a person via localizing joints from images. A lot of practical problems, such as smart environments, human-computer interaction \cite{ye2013survey}, augmented reality \cite{marchand2016pose}, virtual reality, human behaviour analysis and recognition \cite{cheng2015advances}, require the information of human body keypoints. 
Similar to many vision problems, deep learning based methods in the problem of human pose estimation had significant progress in the past few years \cite{xiao2018simple}. While the problem of human pose estimation is in quick developing, we also notice that the network architecture has a tendency of becoming more deep and complex. For example, the winner \cite{chen2018cascaded} of COCO 2017 Keypoint Challenge deploys one ResNet-101 network \cite{he2016deep} for human pose estimation on the top of human detection. The winner \cite{li2019rethinking} of COCO 2018 Keypoint Challenge uses 4 ResNet-50 networks \cite{he2016deep} for keypoint detection refinement. These complex models cause a difficulty in both training and real-world applications. 
In this paper, compared to above mentioned  accuracy-oriented works, we aim to develop a model with high performance as well as speed. 
%In the following, we first analyse the existing pose estimation methods and introduce the application problems faced in the real-world situation. Then 
We propose a new efficient network architecture that strengthened the discrimination power of feature with the limited computational complexity and retain high inference speed.

The key problem for human pose estimation is the keypoint detection. However, one challenging problem is that human left/right keypoints often have very similar appearance. Only relying on detection with local information cannot distinguish the joint type from the left or right. To handle this problem, some previous methods \cite{wei2016convolutional}\cite{newell2016stacked}\cite{chu2017multi} were proposed by increasing the receptive fields to capture global context information to support the discrimination of left/right joint types. In this way, the discrimination of one joint may use the context information of multiple neighbouring body parts. For example, in the single-person cases, authors of \cite{wei2016convolutional} deploy 5 refinement modules stacked on the keypoint detection module, where each refinement module contains 7 convolutional layers with large filters of size 7x7. In \cite{newell2016stacked}, it utilises 8 hourglass-like modules to detect and refine human body keypoints, with each hourglass module consisting 45 convolutional layers. \cite{chu2017multi} adds visual attentional units on the hourglass-like modules to focus on body part regions of interest. In the multi-person scenes, previous methods \cite{cao2017realtime}\cite{newell2017associative}\cite{insafutdinov2016deepercut}\cite{chen2018cascaded}\cite{li2019rethinking} also follow the same network design principals to improve performance.

\begin{figure*}[h!]
\begin{tikzpicture}
  \node (img1) {\includegraphics[width=4.18cm]{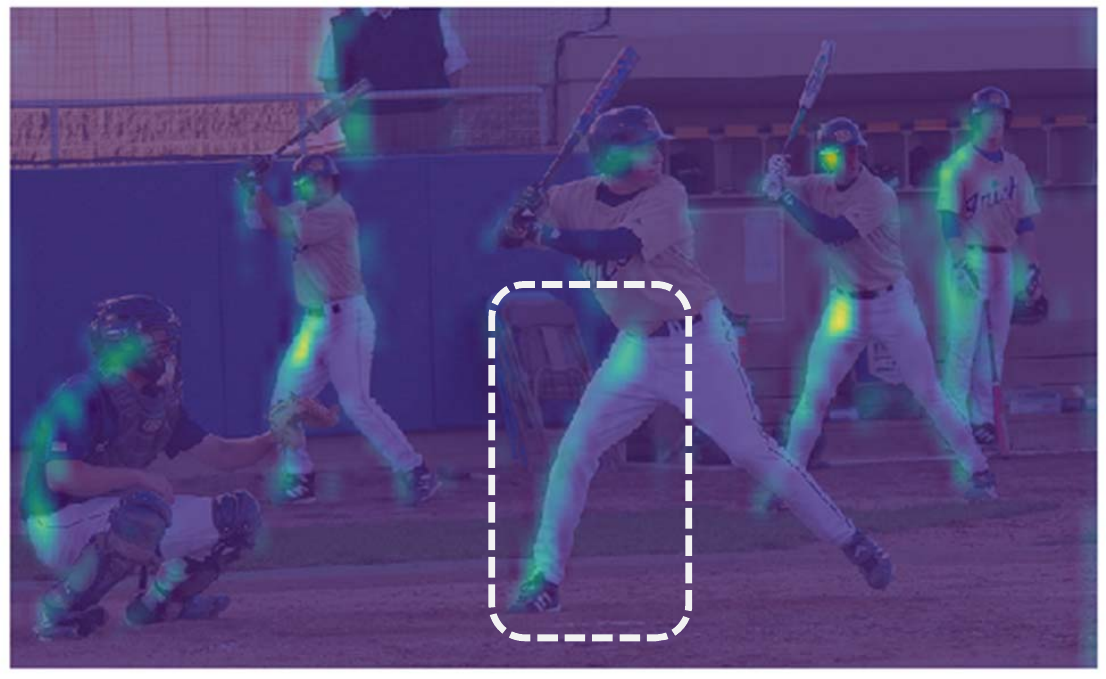} };
  \node[above=of img1, node distance=0cm, yshift=-1cm, font=\color{black}, font=\small] {Conv-1};
  \node[above=of img1, node distance=0cm, xshift=2.2cm, yshift=-0.6cm, font=\color{black}, font=\small] {Joint branch};
  \node[right=of img1, xshift=-1.28cm, yshift=0cm] (img2) {\includegraphics[width=4.18cm]{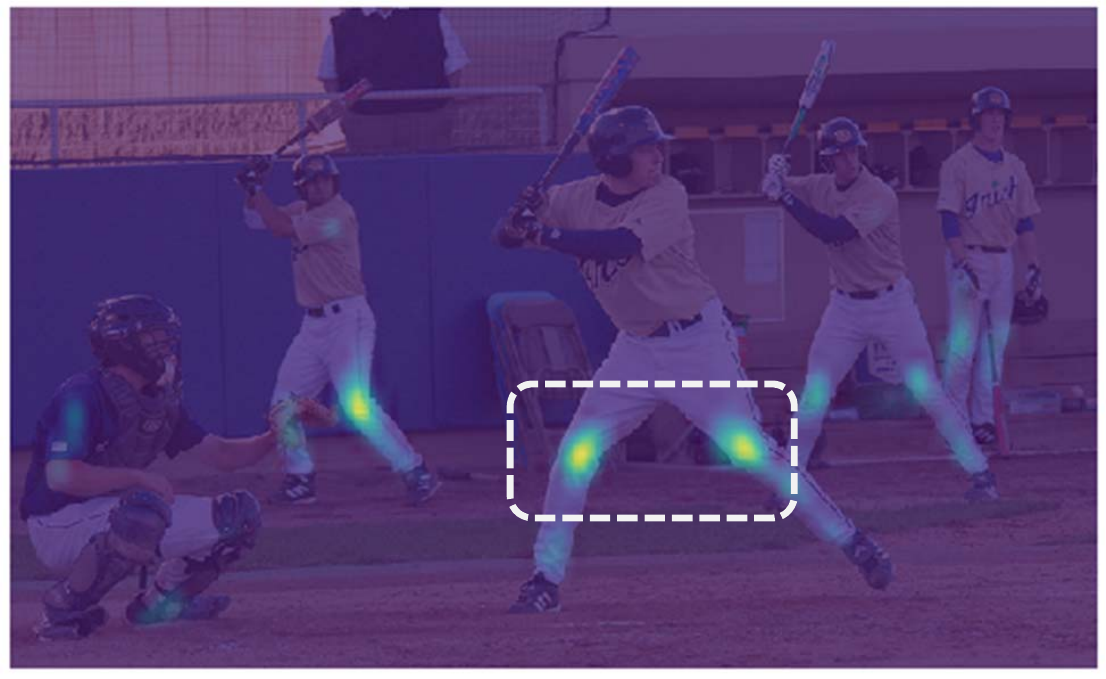}};
  \node[above=of img2, node distance=0cm, yshift=-1cm, font=\color{black}, font=\small] {Conv-4};
  \node[right=of img2, xshift=-0.5cm, yshift=0cm] (img3) {\includegraphics[width=4.18cm]{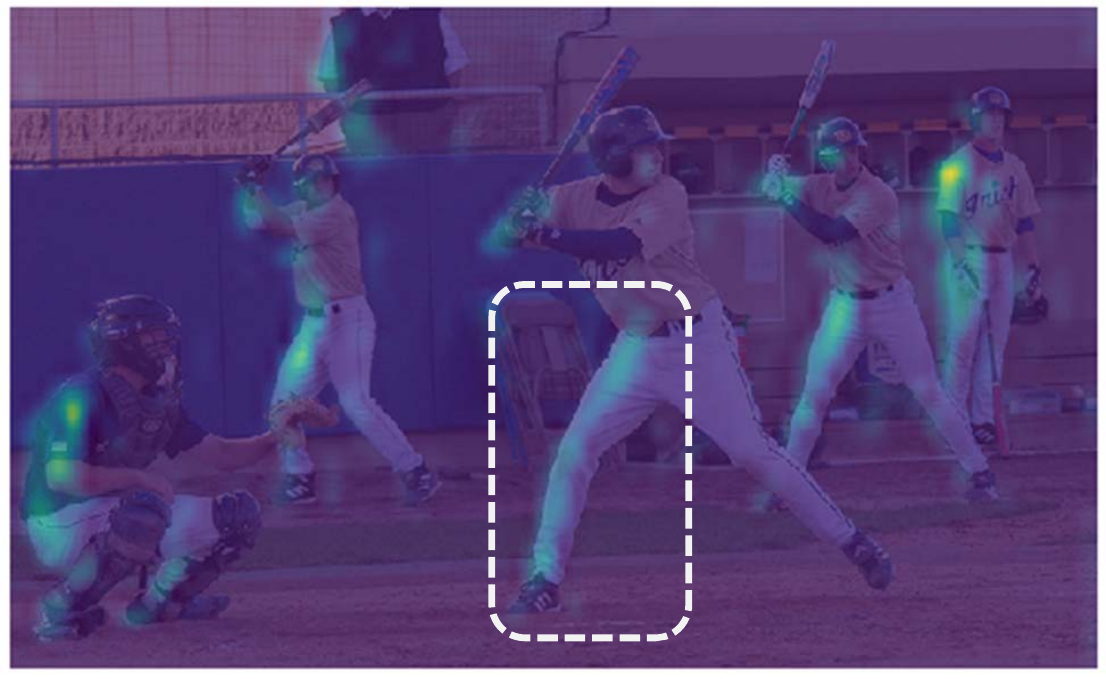}};
  \node[above=of img3, node distance=0cm, yshift=-1cm, font=\color{black}, font=\small] {Conv-1};
  \node[above=of img3, node distance=0cm, xshift=2.2cm, yshift=-0.6cm, font=\color{black}, font=\small] {Limb branch};
  \node[right=of img3, xshift=-1.3cm, yshift=0cm] (img4) {\includegraphics[width=4.18cm]{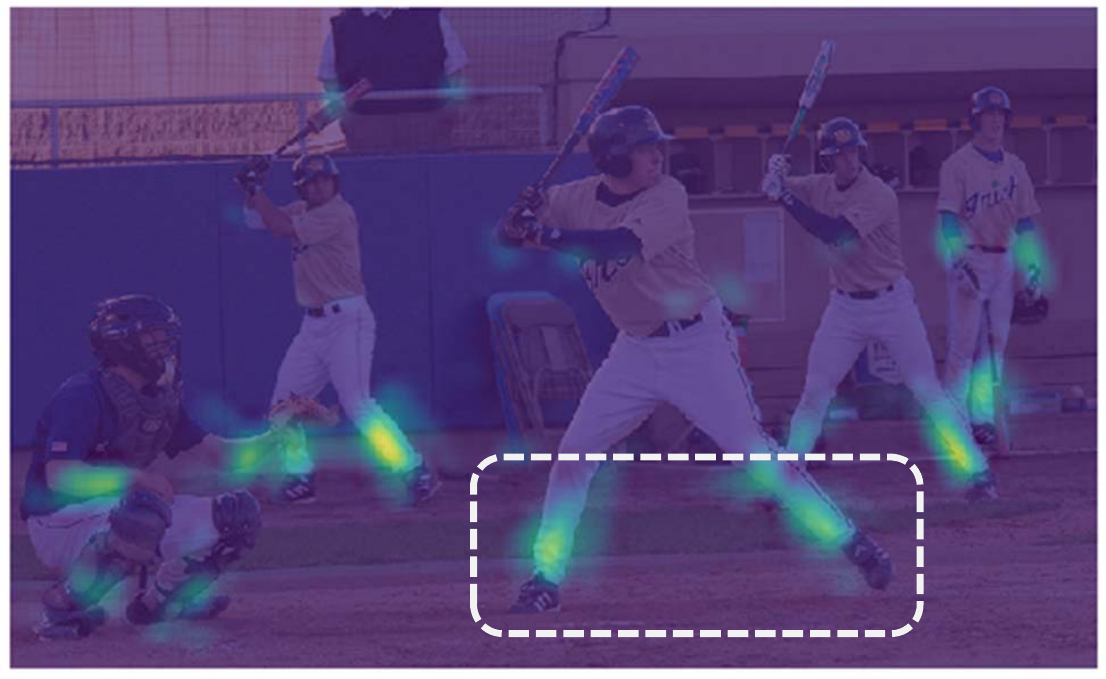}};
  \node[above=of img4, node distance=0cm, yshift=-1cm, font=\color{black}, font=\small] {Conv-4};
\end{tikzpicture}
\caption{Comparisons of feature maps extracted from Conv-1 and Conv-4 (Fig. \ref{fig:netStructure}) of both branches, the features of Conv-1 as a whole naturally own the left/right information, while that of Conv-4 are more localized with the lost left/right information, as highlighted.}%
\label{fig:multi-level}%
\end{figure*}
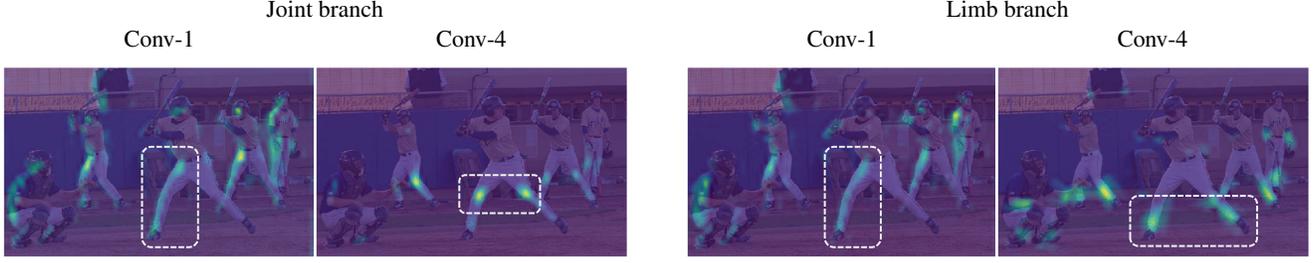

These very deep networks are with high computation load that constrains their application areas due to speed and memory limitations at edge devices \cite{wu2018pocketflow}. We are wondering whether there are some other efficient solutions to support the discrimination of joint type but not increase the cost of computation too much. In our study, we print out the feature maps of detection branches and observe the activation responses of each layer, as shown in Fig. \ref{fig:multi-level}. We find that body part detection module extracted sufficient semantic features. The outputs (Conv-5 in Fig. \ref{fig:netStructure}) of detection layer have very good localization performance. Furthermore, from the first convolutional layer (Conv-1) to the fourth layer (Conv-4), we find that the features in Conv-1 of both branches are abstract. For example, We can see that the upper and lower left legs are activated as a whole region as shown within white-dashed box in Fig. \ref{fig:multi-level}. In contrast, the features in Conv-4 of both branches are more localized. For instance, the activation regions focus on specific body parts. More interestingly, the activated features in Conv-1 as a whole can focus on one left/right type while the features in Conv-4 are highlighted on both left and right types. In other words, the features focus on larger region has better performance than smaller region in discriminating left/right type. This observation inspires us to utilize multiple semantic-wise feature information to strengthen the inference power of the network instead of only increasing the depth of the network.

In this research, we propose a Multi-level Network (MLN) that learns to aggregate multiple semantic-wise features to strengthen the discrimination power of networks. We consider 
aggregated features including lower-level (left/right) information, upper-level (localisation) information, joint-limb level (complementary) information and global-level (context) information. With this architecture, the proposed network achieves comparable accuracy to existing methods with two times more parameters and convolutional layers. Simultaneously, the forward inference speed of the network achieves 42.2 FPS using an input size of 368x432 on one NVIDIA Tesla P40 GPU. This enables real-time high-precision applications in real-world situations.

\section{Multi-level Networks}
\label{sec:Section2}

The objective of pose estimation is to obtain the location of each human body joint in the image. We propose a Multi-level Network (MLN) that can take an image as input, and produce all joints' locations of each person. The MLN architecture consists of the following four modules:

First, body part detection module aggregates lower-level and upper-level features to infer the location of each body part in the image using fully convolutional networks. 
The backbone network is based on the standard VGG-19 \cite{simonyan2014very}, which is used to extract general features. Then the network splits into two branches to extract the specific features of joint and limb, respectively.

\begin{figure}[t!]
 \centering
  \includegraphics[scale=0.95]{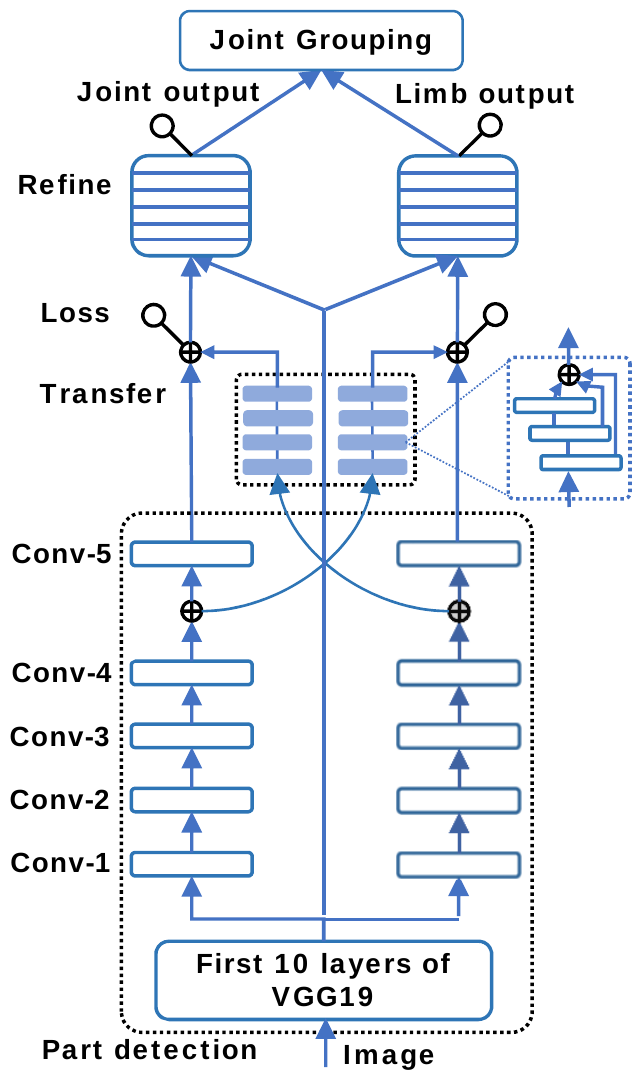}%
  %\caption{Our model contains 4 modules, which are a part detection network, a feature transfer sub-network, a refinement module and joint grouping. Detection module extracts multiple semantic-wise features and concatenate them for enhanced inference. Feature transfer module captures complementary information between joints and limbs. Refinement module combines global context to refine the outputs of detection.}%
  \caption{Network structure of multi-level network (MLN) for keypoint detection using local and global information. It is consisted by four modules and detailed explanations are given in Section 2.1 to 2.4. }
  \label{fig:netStructure}%
\end{figure}

Feature transfer module is then introduced as the second module to cross-transfer the features from one branch to another. The transferred features are then merged with the features of target branch to capture complementary features.
The third module is the refinement module that combines global context information to refine the predictions. The global context is extracted from the bifurcation layer of the backbone network due to this layer extracted abstract semantic features of the whole image.
Finally, joint grouping module distinguishes and assembles joints and limbs for each person. In detail, it associates two neighbouring joints according to the corresponding limb detection. Then it connects each joint pair according to the structure of human body to form a complete body skeleton.
The network structure is shown in Fig. \ref{fig:netStructure}. 

%In the following sections, we give more technical details.

\subsection{Body Part Detection}
\label{ssec:partDetection}

Body part detection uses continuous regression \cite{goodfellow2016deep} since it can provide a smooth transition for the pixels near the annotated joints. A Gaussian function is used to generate joint ground-truth confidence maps. At position $ {\rm p} \in \mathbb{R}^2 $ on the confidence maps, the ground-truth value $ \textbf{J}_i^*(\rm p) $  is defined by:
\begin{equation} 
\textbf{J}_i^*(\rm p)=\underset{c\in[1,k]}{\arg\!\max}\exp\left({-\frac{\|{\rm p}-{\rm p}_{i,c}^*\|_2^2}{\sigma^2}}\right)
\end{equation}
where $\sigma$ is the variance, $ {\rm p}_{i,c}^* \in \mathbb{R}^2 $ is the location of each annotated visible joint, $ c \in [1, k] $, $k$ is the number of visible joints of type $i$.
To facilitate the comparison, PAF \cite{cao2017realtime} is selected as the limb descriptor in our model. The limb ground-truth confidence maps can be defined by:
\begin{equation}
\textbf{L}_j^*({\rm p})=\begin{cases}
\displaystyle
\frac{({\rm p}_{i_1,{\rm h}}^* - {\rm p}_{i_2,{\rm h}}^*)}{\|{\rm p}_{i_1,{\rm h}}^* - {\rm p}_{i_2,{\rm h}}^*\|_2}, & \text{if pixel } {\rm p} \text{ on the limb of person } {\rm h} \\
0 & \text{otherwise}
\end{cases}
\end{equation}
This indicates each of the pixels has a unit vector that points to the next joint. For the pixels are not on the limb region, the vector is zero-valued.

The two loss functions for detection are defined as:
\begin{equation}
e_\textbf{J}=\sum_{i=1}^{m} \sum_{\rm p} {\rm W(p)}\|\textbf{J}_i({\rm p})-\textbf{J}_i^*({\rm p})\|_2^2
\end{equation}
\begin{equation} 
e_\textbf{L}=\sum_{j=1}^{n} \sum_{\rm p} {\rm W(p)}\|\textbf{L}_j({\rm p})-\textbf{L}_j^*({\rm p})\|_2^2
\end{equation}
where W(p) is the binarized mask to ignore unannotated people in the loss computation.

\subsection{Feature Transfer}
\label{ssec:featureTransfer}

We use convolutional operation to implement feature transfer. Feature transfer is calculated according to the equation:
\begin{equation} 
{\rm A}_{\rm b}^{\rm T}=F({\rm A}_{\rm b}\otimes f^{\rm T})
\end{equation}
where $f^{\rm T}$ is the filter bank for feature transfer, $\otimes$ is a convolution operation, and $F$ is the rectified linear unit, ${\rm A}_{\rm b}$ is the feature maps of the branch b, ${\rm A}_{\rm b}^{\rm T}$ is the transferred feature maps.

The transfer sub-network consists of 4 transfer blocks, where each block contains 3 convolutional layers with a kernel size of $3\times3$ and the output features of them are aggregated to strengthen feature propagation. Each convolutional layer has 128 channels with each followed by a ReLU layer except for the last output layer. These 4 refine blocks provide a corresponding receptive field of 100 pixels on the image that can cover the largest distance between adjacent joints on the COCO dataset.
For using the most powerful features, the input of transfer sub-network uses concatenated features in the body part detection module.

\subsection{Refinement Module}
\label{ssec:refineModule}

Refinement module concatenates the heatmaps of joint and limb detection with the global context information and takes them as the input of refinement module. The architecture is shown in Fig. \ref{fig:netStructure}. Both refining branches have the same network configuration, consisting of 7 blocks. Each block uses the same network structure as the transfer block. The stacked convolutional layers increase the receptive field of the network and more context information is captured.

\subsection{Joint Grouping}
\label{ssec:jointGrouping}

The outputs of refine module are the joint and limb confidence maps. Performing non-maximal suppression of 4 neighbourhoods over each score map and the pixels with the largest score in every search are as the corresponding candidate body parts. Then, in the way of greedy search, the matching score of any two pair-wise candidate joints, $\textbf{J}_{i_1}^+$ and $\textbf{J}_{i_2}^+$ from a predefined kinematic chain is computed by the cosine similarity between their line segment and the limb unit vector. More specifically, the matching score is approximated by:
\begin{equation}
s=\sum_{d=1}^{D}\textbf{L}_j^+({\rm p}(d))\frac{(\textbf{J}_{i_1}^+ - \textbf{J}_{i_2}^+)}{\|\textbf{J}_{i_1}^+ - \textbf{J}_{i_2}^+\|_2}
\end{equation}
where $D$ is the total number of line segments.

After computing the matching scores of all the candidate joint pairs, we search for the definite connections of the human skeletons according to the matching score. Search starts from the connection with the highest score and iteratively finding the next connection with the best score until no candidate connections can be found. Finally, we assemble the definite connections that share the same joint to form the complete human skeletons of multiple people.

\begin{table}[!t]
\small
\caption{\label{tab:coco_test-dev}Comparisons of different approaches on the COCO test-dev set. Bold: the best performance. Bold-italic: the performance of our method is better than the baseline. \cite{papandreou2017towards} has not released source code.}
\centering
\begin{tabular}{|p{1.7cm}|p{0.55cm}|p{0.6cm}|p{0.7cm}|p{0.85cm}|p{0.6cm}|p{0.6cm}|}
\hline
Method & FPS & \textbf{AP} & ${\rm AP^0{}^.{}^5}$ & ${\rm AP^0{}^.{}^7{}^5}$ & ${\rm AP^M}$ & ${\rm AP^L}$\\
\hline
G-RMI\cite{papandreou2017towards} & - & 0.605 & 0.822 & 0.662 & \textbf{0.576} & 0.666 \\
Baseline \cite{cao2017realtime} & 20.4 & 0.584 & 0.815 & 0.626 & 0.544 & 0.651 \\
Mask RCNN\cite{he2017mask} & 4.4 & \textbf{0.627} & \textbf{0.870} & \textbf{0.684} & 0.574 & \textbf{0.711} \\
\hline
Our method & \textbf{42.2} & \textbf{\textit{0.584}} & \textbf{\textit{0.821}} & \textbf{\textit{0.626}} & 0.537 & \textbf{\textit{0.658}} \\
\hline
\end{tabular}
\end{table}

\begin{table*}[!t]
\small
\caption{\label{tab:runtime}
Comparisons of our model to other state-of-the-art models. FPS is tested on a single NVIDIA Tesla P40. \cite{he2017mask} corresponds to the configuration of ResNet-50 with feature pyramid network. Our method is 10 times faster than \cite{he2017mask}, and 2 times quicker than the baseline \cite{cao2017realtime}.}
\centering
\begin{tabular}{|p{1.6cm}|p{3.8cm}|p{2.5cm}|p{1.8cm}|p{1.6cm}|p{1.4cm}|p{0.8cm}|}
\hline
Type & Method & Model Size (MB) & \# Parameter & FLOPs & AP & FPS \\
\hline
Top-down & Mask RCNN\cite{he2017mask} & 480.8 & 62.4$\times10^6$ & 536.6$\times10^9$ & \textbf{0.627} & 4.4 \\
\hline
Bottom-up & Baseline \cite{cao2017realtime} & 209.3 & 52.3$\times10^6$ & 159.6$\times10^9$ & 0.584 & 20.4 \\
\hline
Bottom-up & Our method & \textbf{85.2} & \textbf{21.2}$\times$\textbf{10}$^6$ & \textbf{82.5}$\times$\textbf{10}$^9$ & 0.584 & \textbf{42.2} \\
\hline
\end{tabular}
\end{table*}

\section{Experimental Studies}
\label{sec:experiments}

\subsection{MS-COCO Dataset}
\label{ssec:cocoDataset}

MS-COCO 2018 Keypoint Detection dataset \cite{lin2014microsoft} is a very popular dataset in pose estimation. Training and validation sets include 118,287 and 5000 images, respectively, totally containing over 150,000 people. The test-dev set contains roughly 20,000 images. Our models are trained on the training set and evaluated on the validation and test-dev set. The accuracy of test-dev set are provided by the online evaluation server for public comparisons.

Here we mainly use the first 5 metrics of COCO keypoint dataset to evaluate the performance of a model. They are AP (average precision), ${\rm AP^0{}^.{}^5}$, ${\rm AP^0{}^.{}^7{}^5}$, ${\rm AP^M}$, ${\rm AP^L}$, where 0.5 and 0.75 are thresholds, $M$ and $L$ mean middle- and large-sized instance. Notice that ${\rm AP^0{}^.{}^5}$ gives a good accuracy already. AP (averaged across all 10 thresholds) is a stricter metric in which 6 of the matching thresholds exceed $0.70$.

\subsection{Training Details}
\label{ssec:trainingDetails}

During training, stochastic gradient descent is selected with 6e-5 initial learning rate, 0.9 momentum, 0.0005 weight decay, and a mini-batch size of 28. Each sample is augmented with rotating, scaling and flipping. In order to learn a detection confidence within the range of [0, 1] and smooth the training gradients, the pixel values of the cropped patch are normalized by 256 and are subtracted by 0.5. The implementation is built on the open-sourced Caffe framework \cite{jia2014caffe}.

\subsection{Results}
\label{ssec:results}

On the COCO test-dev set, our model achieves the same performance comparing to the baseline model \cite{cao2017realtime} while the speed is 2.1 times faster at the rate of 42.2 FPS. The detailed results are shown in Table \ref{tab:coco_test-dev}. We can see that medium-sized human instances have lower precision than the large-sized people in general. For the ${\rm AP{}^0{}^.{}^5}$ metric our model obtains a very high value of 0.821. 

Total number of parameters of MLN is 21,278,912. In contrast, the baseline model \cite{cao2017realtime} has the parameter number of 52,298,816, which is about 2.5 times more. For the running time of our approach, we record the inference time on a server with one NVIDIA Tesla P40 GPU over 1000 images, which include different numbers of people from 1 to 20. The whole network with 368$\times$432 sized inputs only costs 22.71 ms on average. Group assignment takes 0.2 ms for 2 people and 0.6 ms for 10 people. This can effectively show that MLN has a higher efficiency in both inference speed and performance.

\begin{figure}[!t]
\centering
\begin{tikzpicture}
  \node (img1) {\includegraphics[width=2.85cm]{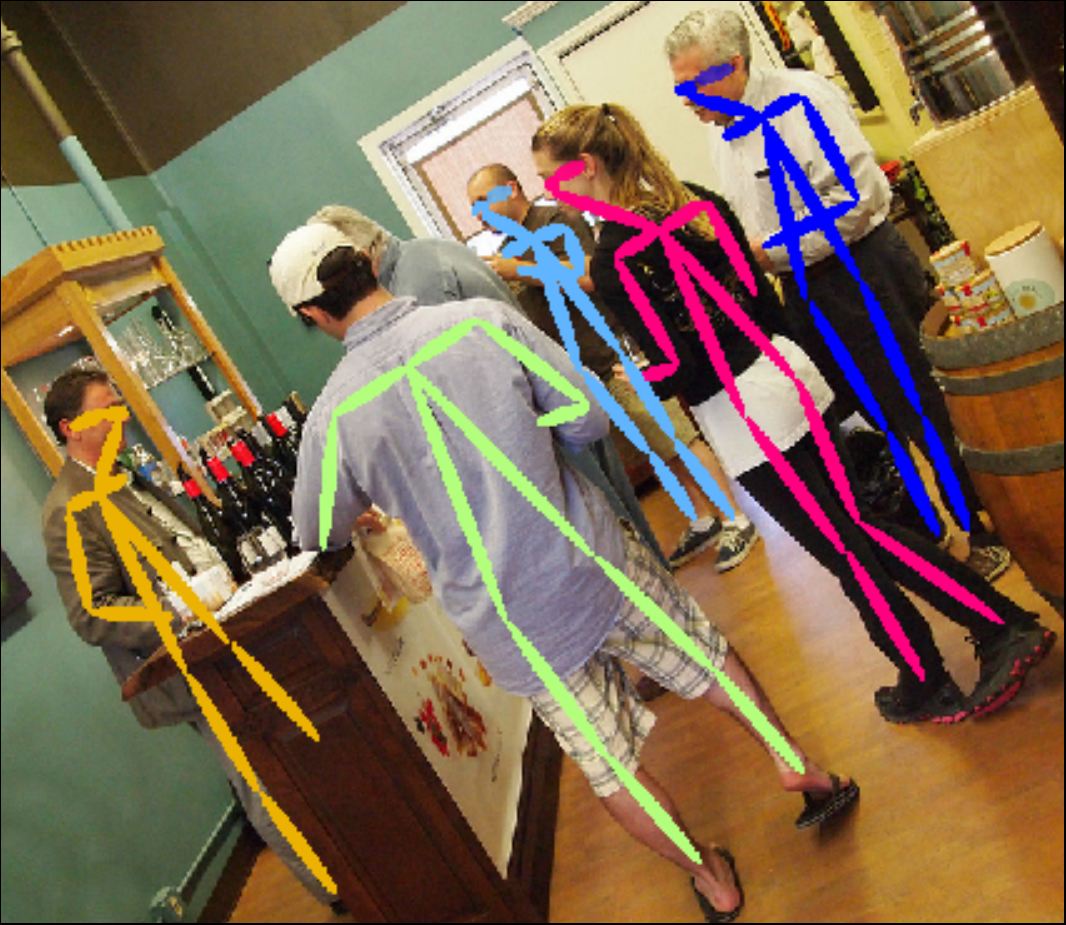} };
  \node[right=of img1, xshift=-1.21cm, yshift=0cm] (img2) {\includegraphics[width=3.7cm]{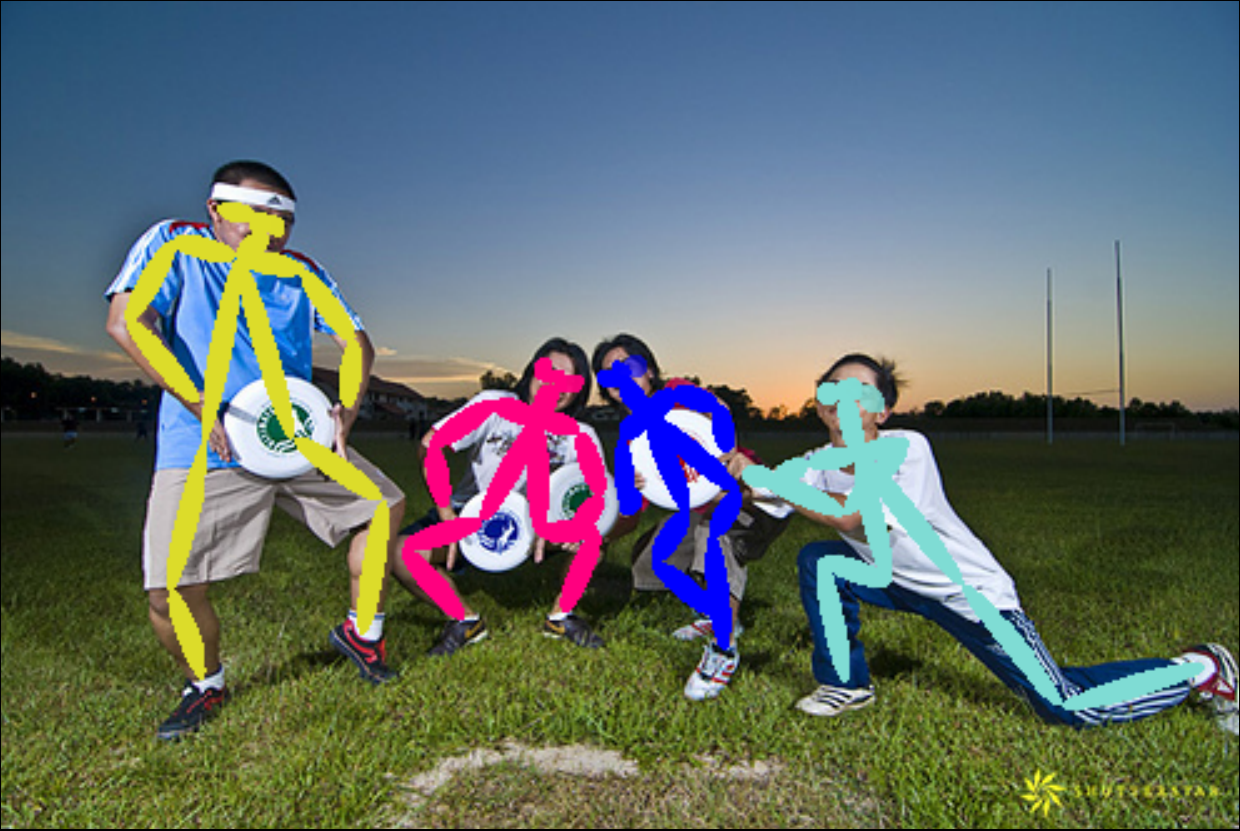} };
  \node[right=of img2, xshift=-1.21cm, yshift=0cm] (img3) {\includegraphics[width=1.65cm]{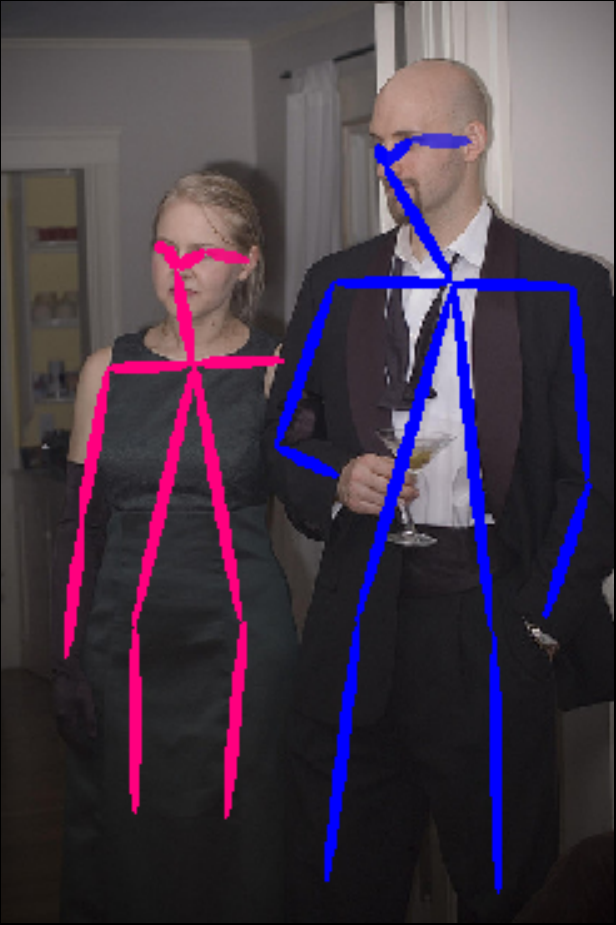} };
  \node[below=of img1, node distance=0cm, xshift=-0.05cm, yshift=1.23cm] (img4) {\includegraphics[width=2.75cm]{22935.pdf}};
  \node[right=of img4, xshift=-1.23cm, yshift=0cm] (img5) {\includegraphics[width=2.75cm]{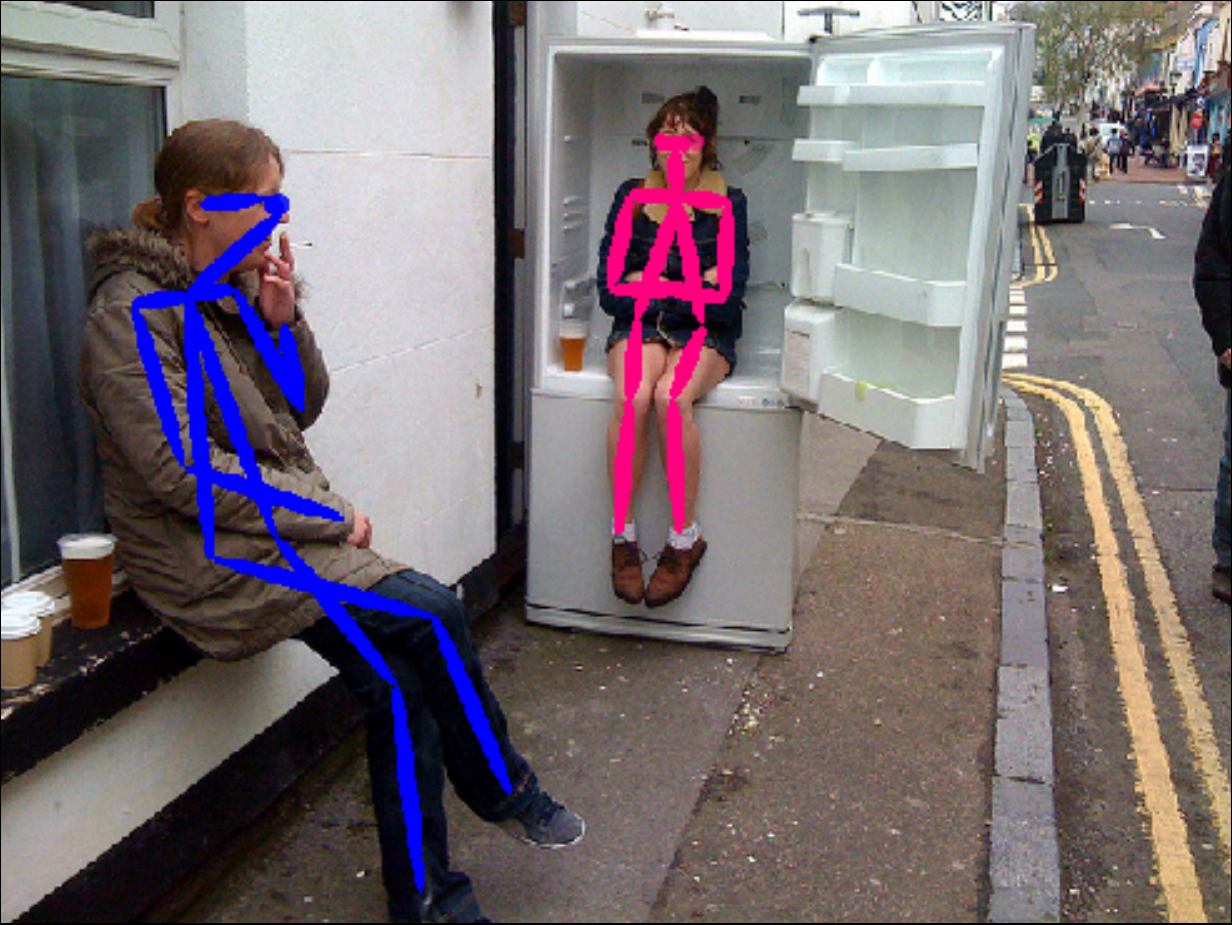}};
  \node[right=of img5, xshift=-1.23cm, yshift=0cm] (img6) {\includegraphics[width=2.75cm]{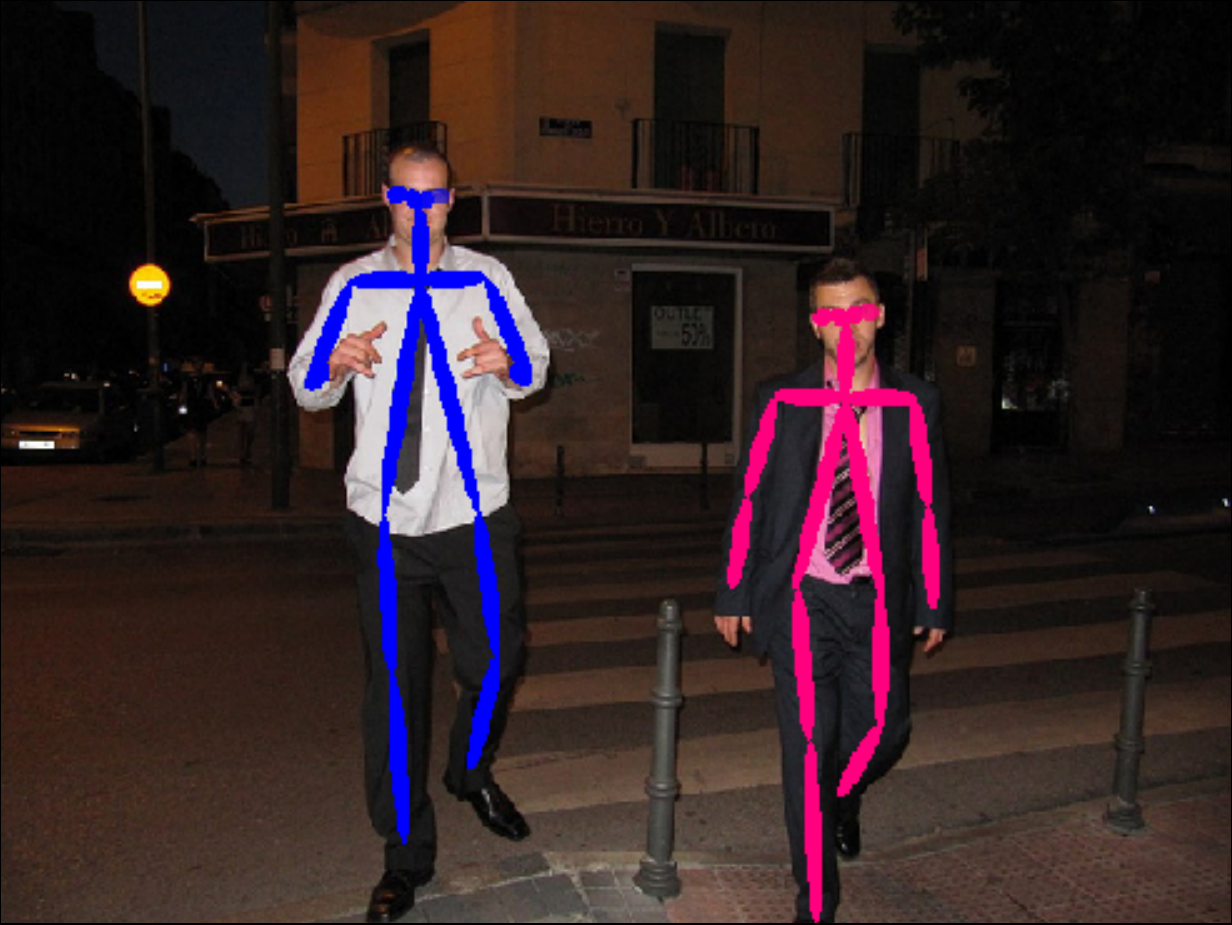}};
\end{tikzpicture}
\caption{Quantitative results of our method on the COCO dataset. Each color corresponds to a human instance.}%
\label{fig:visCOCOResults}%
\end{figure}

We also perform run-time comparisons with the state-of-the-art models on the same environment, except   \cite{papandreou2017towards} who have not released the source code. In Table \ref{tab:runtime}, we compare our method to other two typical top-down and bottom-up methods in terms of computational complexity. Our model is 10 times faster than the top-down one, and 2 times faster than the bottom-up method. In addition, the model size (85.2 MB), number of parameters (21.2$\times$10$^6$) and FLOPs (82.5$\times$10$^9$) of our model are much smaller than other approaches, which facilitates the training and eases the performance requirements of hardware platform.
Finally, Fig. \ref{fig:visCOCOResults} shows some qualitative results, which contains some abnormal cases, such as scale, appearance and viewpoint variation, occlusion and crowding.

\section{Conclusion}
In this work, we have proposed a Multi-level Network that learns to aggregate feature information from lower-level, upper-level, joint-limb level and global-level for discrimination of joint type. Lower-level features embeds left/right information, upper-level features own better localization performance, joint-limb level features provide complementary information, and global-level features add context information. These different semantic-wise features are combined to enhance the inference of joint type while not introducing too much computation cost. The proposed network is validated on the popular COCO Keypoint dataset. Our model achieved comparable state-of-the-art accuracy with inference speed attaining 42.2 FPS, which is between 2 and 10 times faster than existing works.

% As a future direction, improvement is expected by combining other strategies from existing literature, such as the feature pyramid by \cite{yang2017learning}. In addition, various network architectures for structural relation inference tasks in existing works \cite{guler2018densepose, chen2018iterative} could also provide guidance to improve body part detection.

\bibliographystyle{IEEEbib}
\bibliography{strings,refs}

\end{document}